\documentclass[letterpaper]{article} 
\usepackage{aaai25}  
\usepackage{times}  
\usepackage{helvet}  
\usepackage{courier}  
\usepackage[hyphens]{url}  
\usepackage{graphicx} 
\usepackage{natbib}  
\usepackage{caption} 
\usepackage{algorithm}
\usepackage{algorithmic}

\usepackage{newfloat}
\usepackage{listings}
\usepackage[utf8]{inputenc} 
\usepackage[T1]{fontenc}    
\usepackage{url}            
\usepackage{booktabs}       
\usepackage{amsfonts}       
\usepackage{nicefrac}       
\usepackage{microtype}      
\usepackage[dvipsnames]{xcolor}
\usepackage{enumitem}
\usepackage{makecell}
\usepackage{multirow}
\usepackage{subfig}
\usepackage{colortbl}
\usepackage{longtable}
\usepackage{amsmath}
\newtheorem{definition}{Definition}
\usepackage{tcolorbox}

\DeclareCaptionStyle{ruled}{labelfont=normalfont,labelsep=colon,strut=off} 
\floatstyle{ruled}
\newfloat{listing}{tb}{lst}{}
\floatname{listing}{Listing}
\title{Enhancing LLMs via High-Knowledge Data Selection}
\author{
    Feiyu Duan\textsuperscript{\rm 1,5}\footnote{These authors contributed equally to this work and should be considered as co-first authors.}, 
    Xuemiao Zhang\textsuperscript{\rm 2,5*}, 
    Sirui Wang\textsuperscript{\rm 3,5}\thanks{Corresponding authors.}, 
    Haoran Que\textsuperscript{\rm 1}, \\
    Yuqi Liu\textsuperscript{\rm 5}, 
    Wenge Rong\textsuperscript{\rm 4\textdagger},
    Xunliang Cai\textsuperscript{\rm 5}
}
\affiliations{
    \textsuperscript{\rm 1}Sino-French Engineer School, Beihang University, Beijing, China \\
    \textsuperscript{\rm 2}Peking University, Beijing, China \\
    \textsuperscript{\rm 3}Department of Automation, Tsinghua University, Beijing, China \\
    \textsuperscript{\rm 4}School of Computer Science and Engineering, Beihang University, Beijing, China\\
    \textsuperscript{\rm 5}Meituan, Beijing, China \\

    \{duanfeiyu, 2224124, w.rong\}@buaa.edu.cn,
    zhangxuemiao@pku.edu.cn,
    \{liuyuqi, wangsirui, caixunliang\}@meituan.com \\
}

\usepackage{bibentry}

\begin{document}

\maketitle

\begin{abstract}

The performance of Large Language Models (LLMs) is intrinsically linked to the quality of its training data. Although several studies have proposed methods for high-quality data selection, they do not consider the importance of knowledge richness in text corpora. In this paper, we propose a novel and gradient-free \textbf{H}igh-\textbf{K}nowledge \textbf{S}corer (HKS) to select high-quality data from the dimension of knowledge, to alleviate the problem of knowledge scarcity in the pre-trained corpus. We propose a comprehensive multi-domain knowledge element pool and introduce knowledge density and coverage as metrics to assess the knowledge content of the text. Based on this, we propose a comprehensive knowledge scorer to select data with intensive knowledge, which can also be utilized for domain-specific high-knowledge data selection by restricting knowledge elements to the specific domain. We train models on a high-knowledge bilingual dataset, and experimental results demonstrate that our scorer improves the model's performance in knowledge-intensive and general comprehension tasks, and is effective in enhancing both the generic and domain-specific capabilities of the model.

  
\end{abstract}

\section{Introduction}
\label{sec:introduction}


The impressive performance of large language models (LLMs) has been demonstrated in various natural language processing (NLP) tasks \cite{touvron2023llama, openai2023gpt}, yet it is significantly influenced by the quality of the training data \cite{li2023textbooks, xie2024doremi}. Typically, the training data is sourced from extensive text collections such as internet crawls \cite{Patel2020}. However, the quality of such data is frequently inconsistent \cite{kreutzer2022quality}. Therefore, the identification and selection of high-quality data from these vast resources is a critical consideration for achieving optimal model training.

To effectively address this issue, a sub-problem that needs to be solved first is how to define high-quality data. Current methodologies typically adopt one of two approaches: the first involves devising a metric to assess the quality of the text, focusing on attributes such as textual fluency \cite{marion2023less, muennighoff2024scaling}; the second approach involves manually curating a subset from the available corpus, which is considered high-quality based on human experience, to serve as a standard reference, with Wikipedia articles often being chosen for this purpose \cite{xie2024data, engstrom2024dsdm}. Techniques derived from the first approach include strategies to eliminate redundant data \cite{lee2022deduplicating} or employing existing models to compute metrics like perplexity (PPL) \cite{marion2023less} or self-influence scores \cite{thakkar2023self} for the dataset. On the other hand, the latter includes the development of models to assign quality scores \cite{brown2020language} to data or the determination of significance weights \cite{xie2024data} to guide the sampling process.




In practice, the mentioned selection criteria often favor fluent texts that may lack knowledgeable information, as shown in the Appendix Figure 9.
This observation inspires us to propose a novel approach: \textbf{H}igh \textbf{K}nowledge \textbf{S}corer (HKS),  which assesses text quality by detecting the knowledge content of each text sample.
We begin by quantifying the knowledge encapsulated within the texts of training data. Unlike the structured knowledge definitions in \citeauthor{AllenZhu2024PhysicsOL} (\citeyear{AllenZhu2024PhysicsOL}), we simplify knowledge into \textit{knowledge elements} to facilitate quicker knowledge annotation. Following this, we create a multi-domain knowledge element pool consists of 5M knowledge elements from various sources. We employ a multiple pattern matching algorithm to identify all the knowledge elements contained in each text sample and introduce two metrics, \textit{knowledge density} and \textit{knowledge coverage}, to facilitate quantitative analysis. Leveraging these metrics, we propose a comprehensive knowledge scorer to select high-knowledge texts, which are positively correlated with these two metrics.
Benefiting from the categorization of each knowledge element during the creation of the multi-domain knowledge element pool, our approach can select high-knowledge data aligned to the desired domain.






\begin{figure*}[t]
    \centering
    \includegraphics[width=0.85\linewidth]{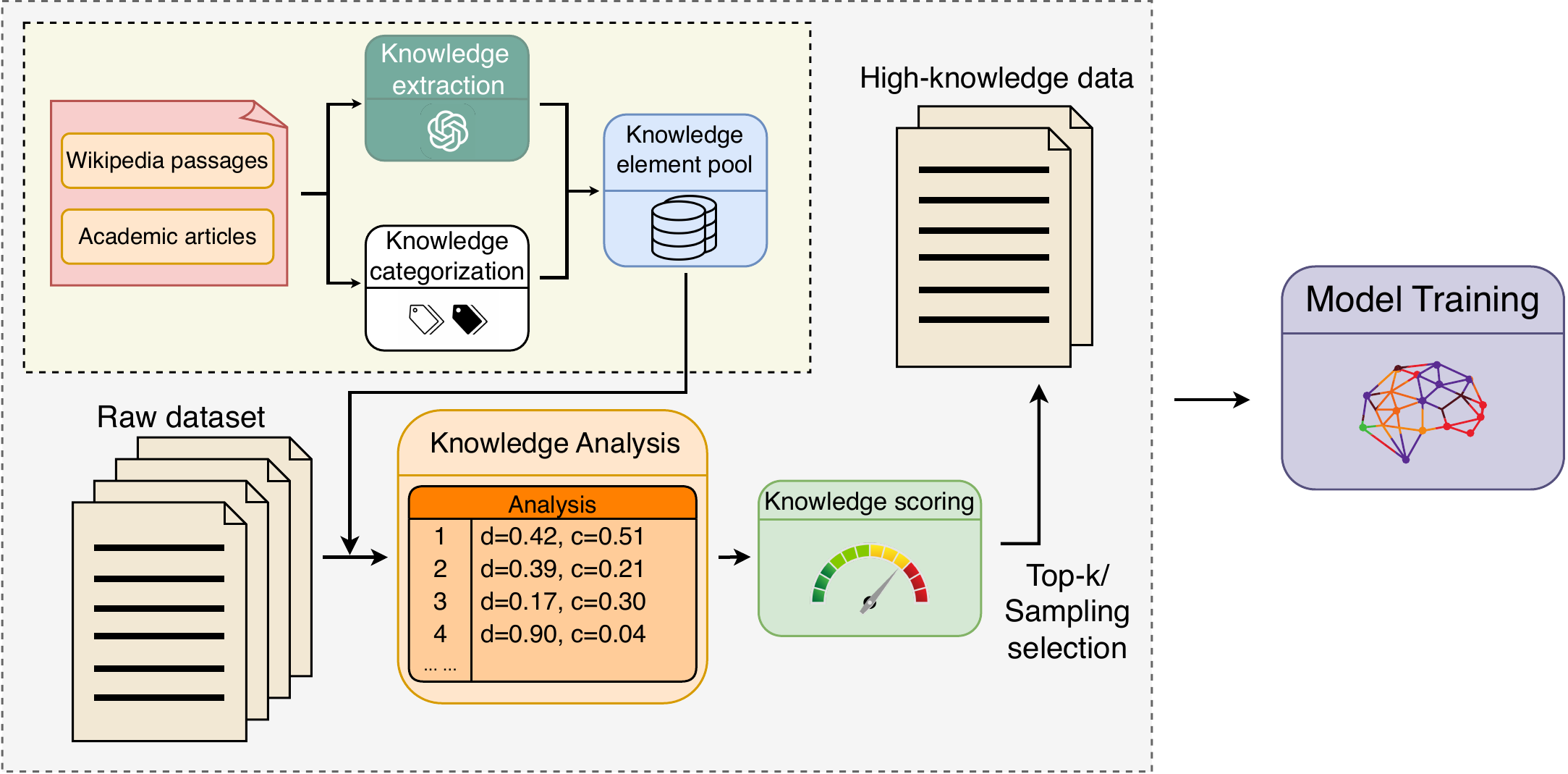}
    \caption{The overall framework of HKS. Our methodology begins with sourcing knowledge from Wikipedia articles and academic literature. We extract knowledge elements, which are categorized into several domains. Each text sample from the raw dataset is then characterized by its knowledge density and coverage, resulting in a knowledge score for each text. Texts with higher scores are identified as high-knowledge data, selected via top-$k$ selection or weighted sampling.}
    \vspace{0mm}
    \label{framework}
\end{figure*}

We train a 1.1B model from scratch on a 20B bilingual dataset. We find that: (1) In comparison to baseline methods, our method excels both in knowledge-intensive tasks and general understanding tasks, with an average improvement of 2.37 pp over random selection. We further validate this conclusion through continual pretraining experiments on Llama-3-8B, observing an average increase of 2.4 pp. (2) When applied to specific domains, our method can enhance the model's performance in the desired domain, with an absolute improvement of up to 2.5 pp compared to baseline.


Our contributions can be summarized as follows:
\begin{itemize}[leftmargin=*]
\item We simplify the definition of knowledge and introduce \textit{knowledge element} to facilitate knowledge parsing of texts, which guides us to establish a multi-domain knowledge element pool. Moreover, we propose knowledge density and coverage as metrics to quantify the knowledge in texts.
\item We propose a novel and gradient-free knowledge scorer for selecting high-knowledge data, which is comprehensive and proportional to knowledge density and coverage. A series of experiments consistently showcase the superior performance of our knowledge scorer.
\item We propose a domain-aware high-knowledge data selection method for domain-specific augmentation. Experimental results demonstrate the effectiveness of our approach.
\end{itemize}


\section{Methodology}
\label{sec:methodology}

\subsection{Overview of Our Approach}

Figure \ref{framework} illustrates the pipeline of our method. Initially, we used Wikipedia and academic literature as knowledge sources. Knowledge elements are extracted and classified using GPT4 \cite{achiam2023gpt} and a BERT-based model. For each text in the pre-training dataset, we enumerate all the included factual knowledge elements, which are then used to determine the two knowledge metrics: density and coverage. A knowledge scorer, proportional to these two metrics, assigns comprehensive scores to each text, which is further considered as data selection criteria.


\begin{figure*}[t]
    \centering
    \includegraphics[width=0.8\linewidth]{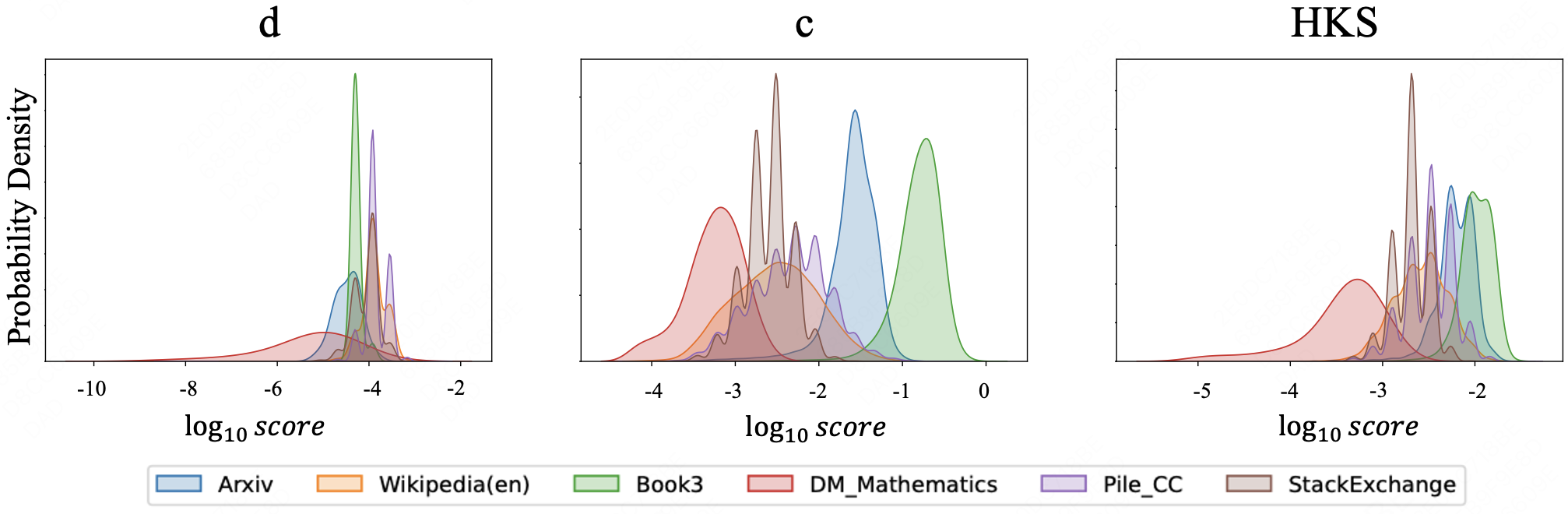}
    \caption{Score density in Pile subsets. There is a noticeable difference in the distribution of knowledge density ($d$) and coverage ($c$) within Pile subsets. $d$ tends to favor samples from Wikipedia, while $c$ tends to favor samples from books and ArXiv.}
    \label{score_density}
\end{figure*}

\subsection{Knowledge Element Pool Construction}



Knowledge is usually represented as structured (name, relation, value) triplets in knowledge graphs \cite{AllenZhu2024PhysicsOL, pan2024unifying}, but parsing triplets across the entire corpus is extremely time-consuming and challenging. To simplify this process, we reduce knowledge triplet to \textit{knowledge element}:


\begin{definition}
A n-gram noun, represented by several tokens $(t_1, t_2, \ldots, t_n)$, can be considered as a \textbf{knowledge element} if it encapsulates or is closely associated with a specific concept, fact, theory, principle, definition, and so forth.\footnote{We only use the names rather than their contents.}
\end{definition}





Based on the above definition, we build a knowledge element pool that covers multiple domains. The knowledge elements are derived from two sources: 1) \textit{Wikipedia documents} 2) \textit{Academic article datasets}. We first add Wikipedia entries and keywords of academic articles to the knowledge element pool, where the academic dataset we use is the OAG dataset \cite{zhang2022oag}. The knowledge elements obtained amount to a total of 20M. While these elements are highly specialized, they lack flexibility. Therefore, we also choose to write instructions to guide GPT4 in extracting knowledge elements contained in the Wikipedia documents\footnote{GPT-4 extracted knowledge elements comprise 13.9\% of the entire pool.}. The prompt is listed in Appendix C.

We categorize the knowledge elements into five domains: \textit{science, society, culture, art, and life}. Given that individual knowledge elements are difficult to label directly, we match related documents for each knowledge element as auxiliary materials for judgment. Then, we manually annotate 10K knowledge elements and take 20\% of the annotated data as the test set. We train a BERT-based \cite{devlin-etal-2019-bert} labeling model with a labeling accuracy of 96.2\%.

After knowledge categorization, we write an additional instruction and use GPT4 to review all the knowledge elements, filtering out those that do not align with the annotated categories or are of poor quality. Additionally, we remove knowledge elements with a string length of less than 2. After deduplication, we finally built a high-quality knowledge element pool containing 5M terms. More detailed construction process and statistics can be found in Appendix E.

\subsection{Knowledge Content Evaluation within the Text}
\label{sec:2.3}




For each text in the training data, we label all the knowledge elements that appear in it\footnote{We use the Aho-Corasick automaton algorithm \cite{pao2010memory} to accelerate our labeling process.}. For quantitative analysis of the pre-training corpus, we establish the following two knowledge metrics.

\begin{definition}
    \label{def:d}
    Given a text sample $x$, $n_k$ is the total number of knowledge elements in the sample, and $n_p$ is the text token length, \textbf{Knowledge Density (d)} is defined as $d(x) = n_k / n_p$.
\end{definition}
    
\begin{definition}
    \label{def:c}
    Given a text sample $x$, $\widetilde{n}_k$ is the total number of non-duplicated knowledge elements in the sample, and $N_k$ is the total number in the full knowledge pool. \textbf{Knowledge Coverage (c)} is defined as $c(x) = \widetilde{n}_k / N_k$.
\end{definition}

Knowledge density is used to quantify the amount of knowledge contained in a text sample. We do not use the total token count of the knowledge elements as $n_k$ to avoid any bias arising from longer knowledge elements. Knowledge coverage serves as an indicator of the diversity of knowledge within a text sample.




We compute two metrics for the texts in the Pile dataset, and present cases in Appendix Figure 9. After separately selecting the top 1M samples from the Pile dataset based on density and coverage, we observe that articles with high knowledge density tend to have an average length of 20,950 tokens, whereas those with high knowledge coverage are typically much shorter, averaging only 811 tokens. This suggests that $d$ and $c$ are two significantly distinct metrics. To examine the observations, we analyze the distribution of $d$ and $c$ across various subsets within the Pile. We select several subsets and draw 10K random samples from each subset. The samples are then divided into buckets based on the minimum and maximum values of the metrics, and we count the number of samples falling into each bucket. Figure \ref{score_density} shows the results, revealing distinct preferences of $d$ and $c$ across different subsets. Notably, $d$ exhibits a preference for subsets such as FreeLaw and Wikipedia, whereas $c$ prefers Arxiv.



\subsection{Knowledge-Based Data Selection} 
\label{sec:knowledge_based_data_selection}

\paragraph{Generic high-knowledge scorer}

According to the aforementioned results, we decide to combine these two orthogonal metrics to create a comprehensive knowledge scorer. Specifically, for a text sample $x$, we materialize the scorer as a scoring function:
\begin{equation}
    score(x) = \phi(d(x), c(x))
    \label{score_function}
\end{equation}


As we have seen in the extraction results (Section \ref{sec:2.3}), these two metrics lead to two completely different extraction results with less entanglement, so we assume that $d$ and $c$ are two variables independent of each other, and we can simplify the function $\phi$ to a product form:
\begin{equation}
    score(x) = f(d(x)) \cdot g(c(x))
\end{equation}

For $f$ and $g$, we give some empirical assumptions:
\begin{enumerate}
    \item $f$ and $g$ should be incremental functions, as we suggest that texts abundant in knowledge information generally yield high scores on knowledge density and coverage.
    \item When the values of $d$ and $c$ are high, their incremental contributions to the overall effect are not expected to be as significant as when their values are low. This implies that the functions $f$ and $g$ do not exhibit upward concavity:

    \begin{equation}
        \frac{\partial^2 f}{\partial d^2(x)}(d(x)) \leq 0,  \frac{\partial^2 g}{\partial c^2(x)}(c(x)) \leq 0
    \end{equation}
\end{enumerate}

We have experimented with various combinations of functions, and the details can be found in Appendix D. 
The scoring formula that we ultimately chose is as follows:

\clearpage

\begin{equation}
    score(x) = d(x) \cdot ln(c(x) + 1)
\end{equation}




The selection cases through our knowledge scorer are displayed in Appendix Figure 9 and 10. Besides, we also analyze the score distribution in Pile. The results in Figure \ref{score_density} show that samples from Book3, Arxiv, and FreeLaw achieve higher HKS scores compared to those from DM Mathematics.

\paragraph{Domain-specific knowledge scorer}


Given that our knowledge elements are categorized, we are able to perform domain-specific knowledge scoring and select high-knowledge data belong to that domain, thereby achieving domain-specific enhancement. Specifically, we constrain our knowledge elements into the target domain $m$, therefore obtaining the knowledge density $d_m$ and coverage $c_m$ for specific domain\footnote{See Appendix E for detailed definition.}.
Similar to the generic knowledge scorer, we can evaluate each text using a domain-specific scoring function for domain $m$:
\begin{equation}
    score_m(x) = d_m(x) \cdot ln(c_m(x) + 1)
    \label{formula:ds_ks}
\end{equation}

\paragraph{Filtering strategies} 
After scoring each text in the pre-training dataset, we adopt two methods for high-knowledge data selection.


\begin{itemize}[leftmargin=*]
    \item \textit{Top-$k$}: 
We select the top-$k$ text samples based on our defined scores, with some selection cases displayed in Appendix Figure 9 and 10. 
It is evident that texts with high scores are more knowledgeable and of higher quality, while samples with low scores contain less knowledge content.


    \item \textit{Sampling}:
In addition to the top-$k$ selection technique, various studies have highlighted the efficacy of sampling methods \cite{sachdeva2024train, wettig2024qurating}. In our research, we employ softmax-based sampling strategies: We treat the normalized score as an importance weight, and apply the softmax function to each sample $x_i$ in the pre-training dataset to calculate sampling probability:
\begin{equation}
    P(x_i) = \frac{exp(\frac{score_i}{\tau})}{\sum_{j}exp(\frac{score_j}{\tau})}
\end{equation}
$\tau$ is the temperature term. We perform sampling without replacement and utilize the Gumbel top-$k$ trick \cite{kool2019stochastic} to facilitate the sampling process. Here we choose $\tau = 2$.
\end{itemize}



\section{Experiments}
\label{sec:experiments}

\subsection{Setups}
\label{sec:experimental_setups}



\paragraph{Dataset}
We utilize the Pile \cite{gao2020pile} and Wudao \cite{yuan2021wudaocorpora} datasets as our pre-training dataset for training a bilingual language model. Pile is an extensive English text corpus that includes 22 diverse subsets, while Wudao consists of Chinese passages collected from web sources. We extract 10B tokens from each, resulting in a 20B token bilingual dataset, which aligns with the compute-optimal amount from previous studies \cite{hoffmann2022training}.


\paragraph{Model and training} 
We train a model of 1.1B parameters, which has the same architecture of Bloom \cite{le2023bloom}.  We train our model in one epoch, with a cosine learning rate scheduler. We use a global batch size of 2048 with gradient accumulation and a max context window length of 2048. We use Megatron framework to train our model in 16 A100 GPUs, with fp16 setting, which needs 21 hours to finish our training. More details can be found in Appendix F.


\paragraph{Baselines}

\begin{figure}[t]
    \centering
    \includegraphics[width=0.9\linewidth]{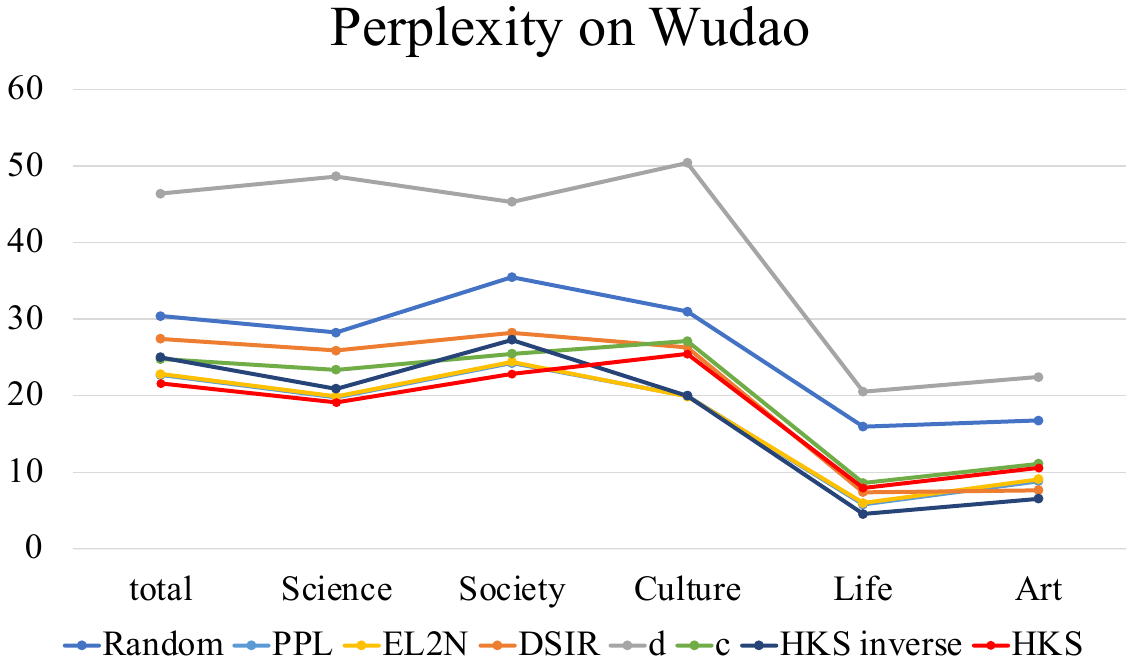}
    \caption{Perplexity evaluation on Wudao validation dataset.}
    \label{ppl_wudao}
\end{figure}



We compare our method with the following baselines: (1) \textit{Random}: Data is selected randomly from each source dataset with uniform probability. (2) \textit{density $d$ and coverage $c$}: We utilize density $d$ and coverage $c$ as the criteria for data selection, respectively. (3) \textit{HKS inverse}: In our ablation studies, we conduct experiments where we select the texts with the lowest scores, as determined by the HKS, to train the model. (4) \textit{Perplexity (PPL)}: In line with the methodology outlined in \citeauthor{marion2023less} (\citeyear{marion2023less}), we employ Bloom 1.1B model \cite{le2023bloom} to calculate the perplexity (PPL) for each data sample. We then retain the top-$k$ samples exhibiting the lowest PPL values. (5) \textit{Error L2-Norm (EL2N)}: Addition to perplexity, we also calculate the error l2-norm for each data sample \cite{marion2023less}. We then retain the top-$k$ samples exhibiting the lowest EL2N values. (6) \textit{Data Selection via Importance Resampling (DSIR)}: Following the methodology outlined in \citeauthor{xie2024data} (\citeyear{xie2024data}), we use bigram representation to compute hash features for each sample. We utilize documents from Wikipedia as the target domain to compute the importance weight.

\begin{table*}[t]
    \centering
    \scalebox{0.95}{
    \begin{tabular}{ll|cc|cc|c}
    \toprule
        \multirow{3}{*}{\textbf{Method}} & ~ & \multicolumn{2}{c}{\textbf{English Tasks}} & \multicolumn{2}{|c|}{\textbf{Chinese Tasks}} & \multirow{2}{*}{\textbf{AVG.}} \\
        \cmidrule{3-6}
        ~ & ~ & \textbf{Knowledge} & \textbf{General} & \textbf{Knowledge} & \textbf{General} & ~ \\
        ~ & ~ & \textbf{intensive} & \textbf{understanding} & \textbf{intensive} & \textbf{understanding} & ~ \\
    \midrule
        Random & ~ & $23.77_{0.06 \  +0.00}$ & $47.56_{0.19 \  +0.00}$ & $25.15_{0.28 \  +0.00}$ & $28.66_{0.14 \  +0.00}$ & $32.49_{0.10 \  +0.00}$ \\
        PPL & ~ & $24.44_{0.07 \  +0.67}$ & $45.92_{0.28 \  -1.64}$ & $24.46_{0.20 \  -0.69}$ & $22.99_{0.06 \  -5.67}$ & $29.41_{0.25 \  -3.08}$ \\
        EL2N & ~ & $24.52_{0.18 \  +0.75}$ & $48.23_{0.22 \  +0.67}$ & $26.11_{0.10 \  +0.96}$ & $24.34_{0.10 \  -4.32}$ & $30.84_{0.23 \  -1.65}$ \\
        DSIR & ~ & $19.26_{0.20 \  -4.51}$ & $48.12_{0.03 \  +0.56}$ & $25.61_{0.07 \  +0.46}$ & $23.97_{0.23 \  -4.69}$ & $29.75_{0.02 \  -2.74}$ \\
    \midrule
        \multirow{2}{*}{$d$} & \multicolumn{1}{|l|}{Sampling} & $24.34_{0.09 \  +0.57}$ & $48.04_{0.27 \  +0.48}$ & $25.46_{0.27 \  +0.31}$ & $26.34_{0.15 \  -2.32}$ & $31.64_{0.14 \  -0.85}$ \\
        ~ & \multicolumn{1}{|l|}{Top-$k$} & $21.72_{0.21 \  -2.05}$ & $43.75_{0.20 \  -3.81}$ & $24.79_{0.01 \  -0.36}$ & $24.39_{0.23 \  -4.27}$ & $29.11_{0.11 \  -3.38}$ \\
    \midrule
        \multirow{2}{*}{$c$} & \multicolumn{1}{|l|}{Sampling} & $23.70_{0.29 \  -0.07}$ & $43.08_{0.29 \  -4.48}$ & $26.75_{0.14 \  +1.60}$ & $26.71_{0.29 \  -1.95}$ & $30.55_{0.18 \  -1.94}$ \\
        ~ & \multicolumn{1}{|l|}{Top-$k$} & $25.27_{0.29 \  +1.50}$ & $43.05_{0.05 \  -4.51}$ & $26.21_{0.01 \  +1.06}$ & $27.45_{0.27 \  -1.21}$ & $31.08_{0.20 \  -1.41}$ \\
    \midrule
        \multirow{2}{*}{HKS} & \multicolumn{1}{|l|}{Inverse} & $20.29_{0.00 \  -3.48}$ & $46.12_{0.11 \  -1.44}$ & $26.55_{0.25 \  +1.40}$ & $25.21_{0.21 \  -3.45}$ & $30.08_{0.28 \  -2.41}$ \\
        ~ & \multicolumn{1}{|l|}{Sampling} & $24.89_{0.11 \  +1.12}$ & $43.61_{0.22 \  -3.95}$ & $26.74_{0.19 \  +1.59}$ & $26.98_{0.24 \  -1.68}$ & $31.00_{0.02 \  -1.49}$ \\
        ~ & \multicolumn{1}{|l|}{Top-$k$} & $\textbf{26.32}_{0.11 \  +2.55}$ & $\textbf{48.93}_{0.24 \  +1.37}$ & $\textbf{27.37}_{0.16 \  +2.22}$ & $\textbf{32.00}_{0.26 \  +3.34}$ & $\textbf{35.07}_{0.26 \  +2.58}$ \\
    \bottomrule
    \end{tabular}
    }
    \caption{Few-shot results of downstream tasks. Bold indicates the best result in each column. All results are averaged over 3 seeds, with standard deviations indicated in subscripts. Signed numbers indicate the difference in scores from the random baseline.}
    \label{main_results}
\end{table*}

\paragraph{Benchmarks and metrics} We train all the models three times using different random seeds and report the average results. We conduct a holistic assessment of all the methods:


\begin{itemize}[leftmargin=*]
    \item We measure perplexity on both Pile and Wudao datasets. For the Pile dataset, we extract 10K samples from each subset to serve as a validation dataset, ensuring these samples are not encountered during the training process. Since the Wudao dataset does not have a predefined subset split, we divide it according to categories of the included knowledge elements. We then apply the same validation process as with the Pile dataset, extracting samples for evaluation.
    \item We assess downstream task performance using in-context learning \cite{dong2022survey}. For knowledge-intensive tasks, we conduct evaluations on ARC-C \cite{bhakthavatsalam2021think}, OpenBookQA \cite{Mihaylov2018CanAS}, MMLU \cite{hendrycks2020measuring}, CMMLU \cite{li2024cmmlu}, and C-Eval \cite{huang2023ceval}. Additionally, we evaluate the model's general understanding capabilities on a range of tests, including RTE \cite{wang2018glue}, BBH \cite{suzgun2023challenging}, WiC \cite{pilehvar2019wic}, COPA \cite{roemmele2011choice}, BoolQ \cite{clark2019boolq} and sub-tasks derived from CLUE \cite{xu2020clue} and FewCLUE \cite{xu2021fewclue}. These test sets encompass both English and Chinese languages. We summarize our test results on \textit{knowledge intensive} and \textit{general understanding} tasks, more details can be found in Appendix G.
\end{itemize}

\subsection{Main Results}



Table \ref{main_results} details the results of our main tests. 
(1) Firstly, we can find that our HKS outperforms the baseline models in most tasks, demonstrating that high-knowledge data can improve the performance of LLMs. Notably, HKS exhibits superior performance relative to PPL, EL2N, and random sampling in terms of average score, with 2.37 pp improvement compared to random sampling, which shows the efficacy of our knowledge scorer. Furthermore, HKS outperforms DSIR in knowledge-intensive tasks, achieving a 0.81 pp improvement. While DSIR uses Wikipedia passages as its target domain, which are rich in knowledge, HKS demonstrates greater efficacy in selecting high-knowledge data.
(2) In addition, the performance of the $d$ and $c$ baselines is inferior to that of HKS, which underscores the importance of integrating density and coverage into a comprehensive knowledge scorer. On the other hand, inverse selection of HKS yields lower results than a random baseline, further affirming the efficacy of our approach from a different perspective. 

We report the results of the two filtering methods, top-$k$ and sampling. The results show that except for $d$, top-$k$ is better than sampling, which indicates that in the dimension of knowledge, the top-ranked data do have higher quality than lower-ranked, indirectly reflecting that our knowledge scorer can accurately identify the high-quality data in the dataset.


\begin{figure*}[t]
    \centering
    \includegraphics[width=0.9\linewidth]{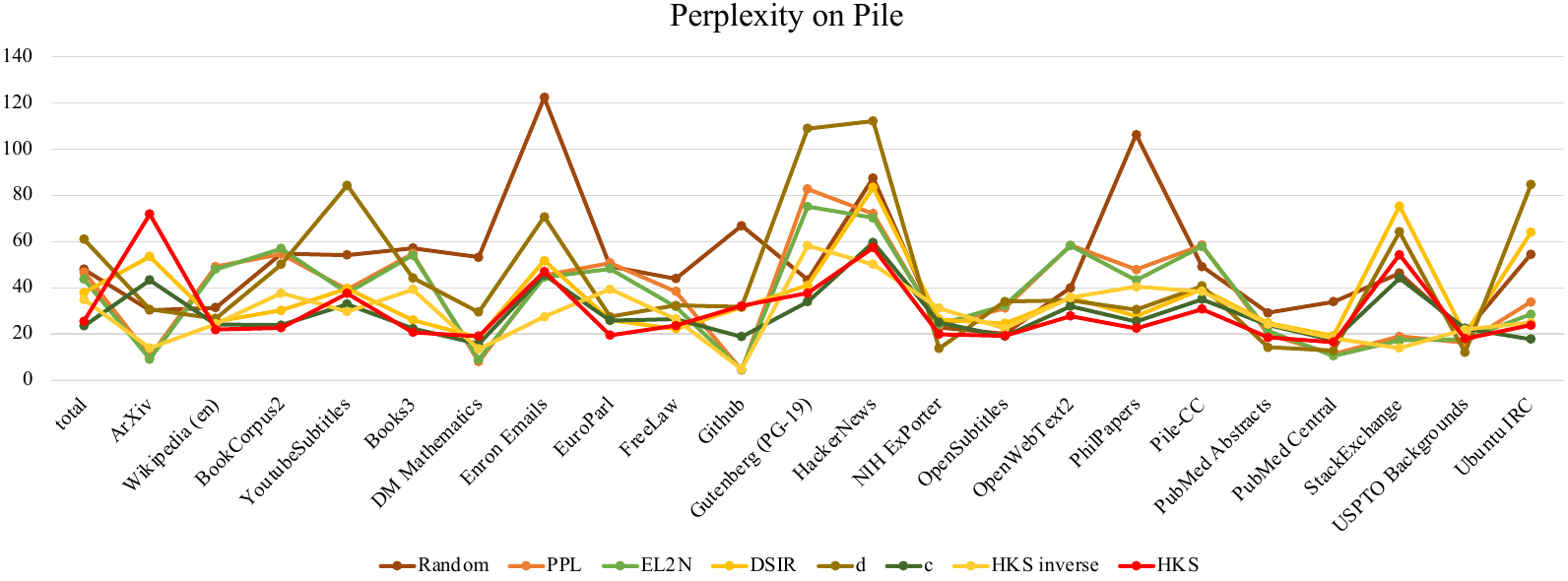}
    \caption{Perplexity evaluation on Pile validation dataset.}
    \vspace{0mm}
    \label{ppl_pile}
\end{figure*}

We also present the performance of our model on perplexity in Figures \ref{ppl_wudao} and \ref{ppl_pile}. 
On the Pile dataset, our perplexity is marginally higher than that of the $c$ baseline; however, HKS outperforms the $c$ baseline regarding downstream task scores, indicating that perplexity may not be strictly positively correlated with downstream task performance. In the context of specific subsets, our model records a lower perplexity on the Wikipedia and book corpora, which are generally regarded as high-quality sources abundant in knowledge.





\subsection{Analysis}

\subsubsection{Extending to Larger Scale}

To conduct larger-scale verification, we select 100B tokens from the Pile and Wudao datasets to continue training on Llama-3-8B \cite{dubey2024llama}. The results are reported in Table \ref{ct_results}. Compared to random selection, our approach results in improvements of 0.9 pp, 3.8 pp, and 2.7 pp on MMLU, CMMLU, and C-Eval, respectively. Furthermore, in comparison to the original Llama-3-8B, our model, post-continual training, exhibits a significant enhancement in knowledge-intensive tasks. The results demonstrate that HKS can be effectively applied to larger datasets and models.

\begin{table}[t]
    \centering
    \scalebox{0.9}{
        \begin{tabular}{l|c|ccc}
        \toprule
            Method & Tokens & MMLU & CMMLU & CEVAL \\ 
        \midrule
            Llama-3-8B & / & 65.8 & 51.5 & 50.8 \\ 
            + Random & 100B & 65.9 & 53.7 & 52.5 \\ 
            + HKS & 100B & \textbf{66.4} & \textbf{57.5} & \textbf{55.2} \\ 
        \bottomrule
        \end{tabular}
    }
    \caption{Comparison of continual pretrained models.}
    \label{ct_results}
\end{table}

\subsubsection{Higher Knowledge Richness in Data Benefits Model Performance}
\label{sec:higher_knowledge}



To further investigate the effects of knowledge richness in data on our final model performance, we sort the texts in Pile and Wudao from high to low according to their HKS scores, and then employ a top-$k$ strategy to select the highest-scoring portion of the data so that the total length of their tokens is 10B, respectively. The score of the lowest-scoring sample in the selected data is determined to be the threshold for the division of high-knowledge data and low-knowledge data. Then we define $\alpha = \frac{N_h}{N_h + N_l}$, where $N_h$ and $N_l$ denote the number of tokens of high and low-knowledge data, respectively. According to the $\alpha$, we perform uniform sampling from both the high and low knowledge portions. The sampled subsets are then merged to form a 20B token dataset, which is utilized for training our model.


\begin{table}[t]
    \centering
    \scalebox{0.95}{
    \begin{tabular}{l|ccccc}
    \toprule
        \textbf{$\alpha$} & \textbf{MMLU} & \textbf{CMMLU} & \textbf{C-Eval} & \textbf{BBH} & \textbf{AVG.} \\ 
        \midrule
        
        1.00 & \textbf{27.91} & 27.85 & \textbf{26.89} & \textbf{29.66} & \textbf{28.08} \\ 
        0.75 & 27.07 & \textbf{27.90} & 26.60 & 26.38 & 26.67 \\ 
        0.50 & 25.78 & 26.63 & 25.47 & 26.98 & 26.53 \\ 
        0.25 & 26.67 & 26.11 & 25.50 & 26.62 & 26.23 \\ 
        0.00 & 25.90 & 25.25 & 24.85 & 27.17 & 25.79 \\ 
    \bottomrule
    \end{tabular}
    }
    \caption{Impact of data quality on model performance. We train models on the merged datasets, varying the proportion of high-knowledge data $\alpha$.}
    \label{high_low_quality}
\end{table}

The results outlined in Table \ref{high_low_quality} demonstrate a trend of diminishing average performance of both knowledge-intensive (MMLU, CMMLU, C-Eval) and reasoning (BBH) tasks, as we move from a dataset consisting entirely of high-knowledge data ($\alpha=1.00$) to the one that is solely comprised of low-knowledge data ($\alpha=0.00$). This indicates that the knowledge content of the data not only affects the model's ability to memorize knowledge but is also potentially linked to the model's reasoning abilities. In addition, each benchmark responds differently to changes in the knowledge content of data. For instance, the results of CMMLU show an increase in performance when the dataset includes a mixture of high and low-knowledge data ($\alpha=0.75$), whereas the results of MMLU, C-Eval, and BBH tend to perform better with higher proportions of high-knowledge data.

\subsubsection{Domain-Specific Enhancement Results}



We explore the application of the HKS model to specific domains, taking the enhancement of the science domain as an illustrative example. We use the Equation \ref{formula:ds_ks} to cherry-pick data rich in scientific knowledge. Similar to Section \ref{sec:higher_knowledge}, we distinguish between high and low-scientific knowledge data by a score threshold at 10B. Subsequently, we define $\beta = \frac{N_{hs}}{N_{hs} + N_{ls}}$ where $N_{hs}$ and $N_{ls}$ denote the token count from high and low-scientific knowledge data, respectively. We follow $\beta$ to perform uniform sampling across these two distinct parts and finally mixed into 20B token training data. We also categorize the questions in MMLU and CMMLU into scientifically relevant and irrelevant sections\footnote{Questions that fall under the STEM category are considered as \textit{science-related}.}, and evaluate the model's performance in these different partitions.

\begin{table}[t]
    \centering
    \scalebox{0.95}{
        \begin{tabular}{l|cc|cc}
        \toprule
            \multirow{2}{*}{\textbf{$\beta$}} & \multicolumn{2}{c|}{\textbf{MMLU}} & \multicolumn{2}{c}{\textbf{CMMLU}}  \\ 
        \cmidrule{2-5}
            \multirow{2}{*}{} & \textbf{Science} & \textbf{Others} & \textbf{Science} & \textbf{Others} \\ 
        \midrule
            1.00 & \textbf{27.29} & 24.16 & \textbf{28.88} & 24.52  \\ 
            0.75 & 26.63 & 24.18 & 28.65 & 25.36  \\ 
            0.50 & 25.56 & 24.97 & 27.26 & 25.10 \\ 
            0.25 & 25.62 & 23.81 & 26.38 & 25.15 \\ 
            0.00 & 22.69 & 23.53 & 26.22 & 25.10 \\ 
        \midrule
            Random & 26.52 & 24.78 & 26.39 & 25.12 \\
        \bottomrule
        \end{tabular}
    }
    \caption{We carry out experiments focusing on scientific knowledge enhancement, where $\beta$ signifies the proportion of high scientific knowledge data within the entire dataset.}
    \label{s/ns}
\end{table}


The results are summarized in Table \ref{s/ns}. We can find that: (1) Our domain-specific HKS is effective in selecting high-knowledge data in the desired domain. Within the \textit{Science} category of the MMLU and CMMLU, the optimal performance is attained when the value of $\beta$ is set to 1.00, with improvements of 0.77 pp and 2.49 pp compared to the random baseline, respectively. The results strongly imply that there is a direct correlation between the amount of HKS-selected domain-specific data and the final result within that domain, indirectly underscoring the effectiveness of domain-specific enhancement through our knowledge scorer. (2) Conversely, the most suboptimal performance across all categories is observed when $\beta$ is set to 0.00. This situation is marked by the absence of high-scientific knowledge data in the training dataset, indicating that such data significantly contribute to the overall performance of the model.




\subsubsection{HKS Achieves Superior Cost Efficiency}
\label{sec:method_costs}




\begin{figure}[t]
    \centering
    \begin{minipage}{0.50\linewidth}
        \centering
        \includegraphics[width=1.0\linewidth]{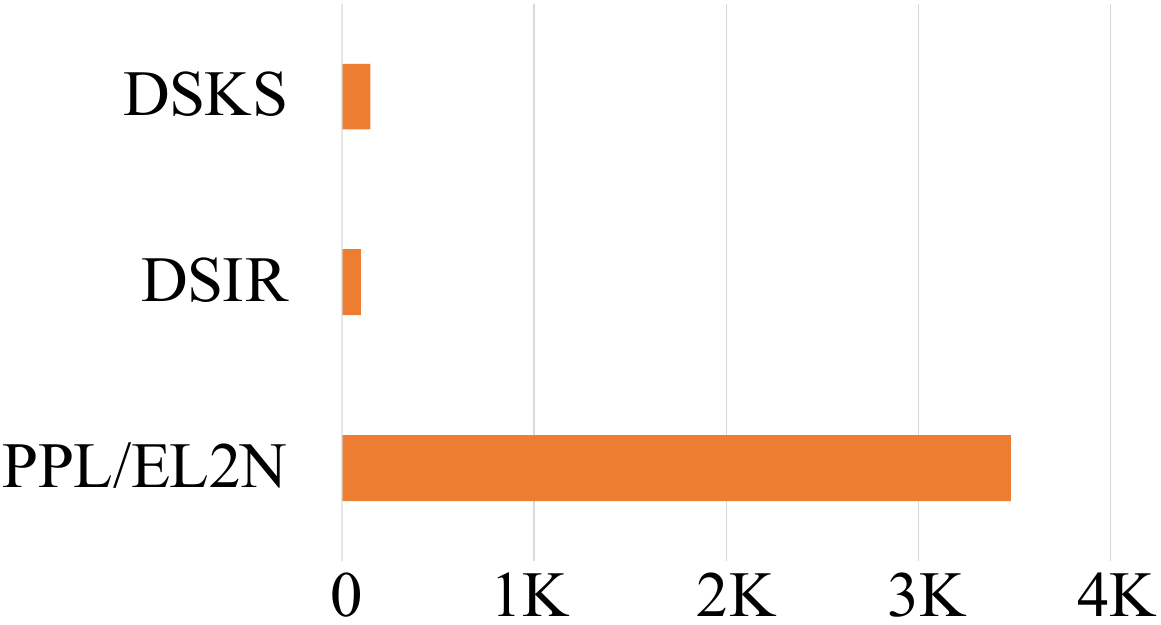}
        \caption{We compare the costs of the various approaches based on cloud server rental fees.}
        \label{costs}
    \end{minipage}
    \quad
    \begin{minipage}{0.41\linewidth}
        \centering
        \includegraphics[width=1.0\linewidth]{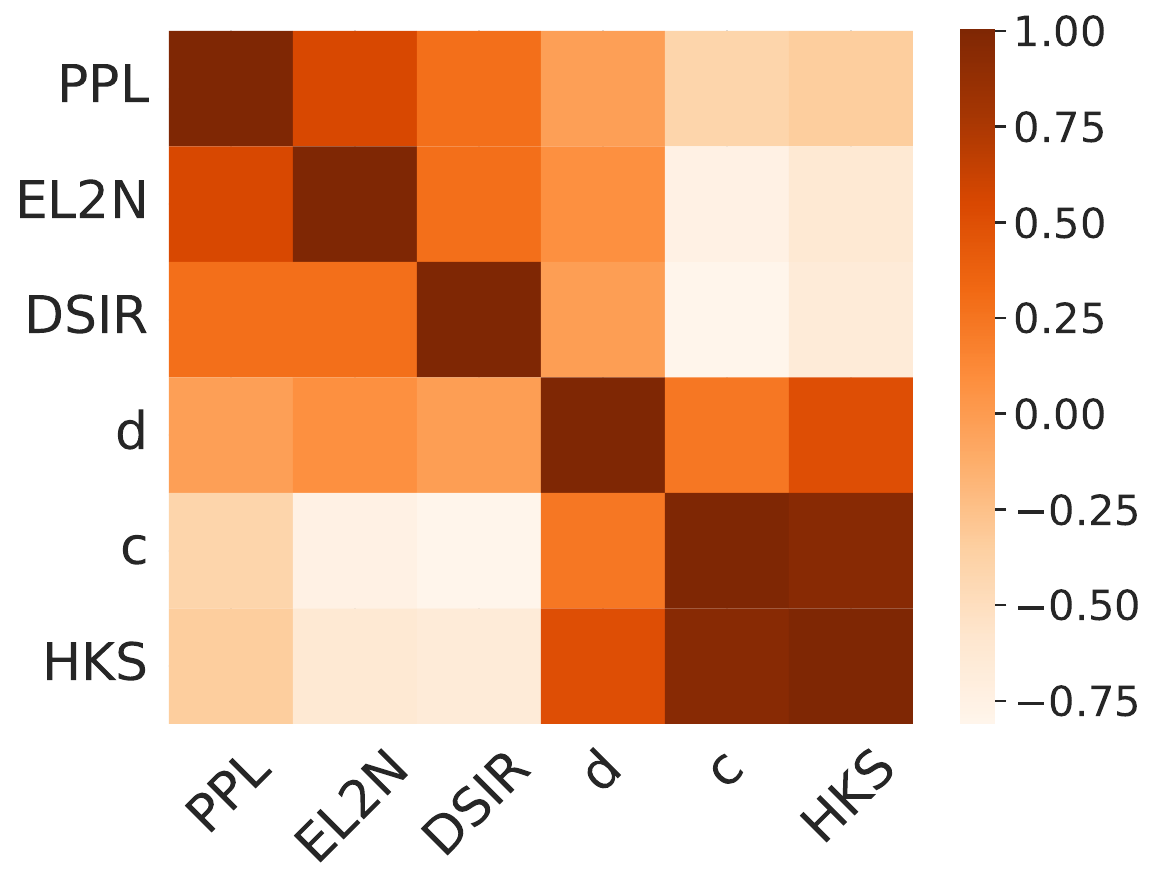}
        \caption{We use Spearman's rank correlation to assess the relation between various methods.}
        \label{corr_coef}
    \end{minipage}
    \vspace{0.1mm}
\end{figure}

Our methodology employs a gradient-free knowledge scorer, which enables our scoring program to run efficiently on a CPU machine. This offers significant cost and time advantages compared to methods such as Perplexity \cite{marion2023less} or model-based scoring \cite{li2023textbooks}. To facilitate a more equitable comparison, we consult the rental rates for CPU and GPU servers on Azure\footnote{https://azure.microsoft.com/en-us/pricing/details/machine-learning/} and present the costs of the different methods in Figure \ref{costs}\footnote{The cost is determined by multiplying the number of CPUs/GPUs, the execution time, and the hourly rate per device.}. Our method incurs considerably lower expenses than the PPL/EL2N, and although it is marginally more costly than DSIR, it delivers superior results. 



\subsubsection{Score Correlation Analysis}

To investigate the correlation between our method and the baseline methods, we extract a portion of the training data for analysis.
We randomly sample 500K texts from the Pile dataset and label this subset with perplexity, error L2-norm, DSIR, $d$, $c$, and HKS scores. Then we calculate Spearman's rank correlation coefficient \cite{gauthier2001detecting} between these scoring methods pairwise. 

From the results depicted in Figure \ref{corr_coef}, we can find out that: 
(1) The correlations between the HKS and baseline methods are remarkably low, trending towards a negative correlation. Perplexity is frequently employed as a metric for evaluating linguistic fluency, suggesting that high-knowledge and fluency do not necessarily correlate strongly. Nevertheless, the HKS still delivers superior performance, indicating that compared to text fluency, high-knowledge data are more beneficial for model training. 
(2) There is a low correlation between $d$ and $c$, indicating that these dimensions are relatively orthogonal to each other, which is consistent with our observation in Section \ref{sec:2.3}.
(3) HKS scores show a strong correlation with both $d$ and $c$, indicating that our scoring function effectively synthesizes metrics along these two dimensions.

\section{Related Works}



\paragraph{High-quality data selection}
Research on high-quality training data selection falls into two approaches: metric-based selection and reference data-based selection. The former typically employs a manually defined metric to assess data quality.
Rule-based methods \cite{rae2021scaling, dodge2021documenting, cai2024internlm2} and deduplication \cite{lee2022deduplicating, touvron2023llama} are two widely used approaches, which are straightforward but may not comprehensively evaluate data quality. Model-based methods leverage language models to derive data attributes, such as perplexity \cite{marion2023less}, self-influence \cite{thakkar2023self}, or density \cite{sachdeva2024train}. Several researchers have also explored directly querying LLMs to give data quality scores \cite{sachdeva2024train, wettig2024qurating}. While these methods may enhance the model's performance, they have not taken into account the data from the perspective of knowledge content.

Reference data-based selection involves pre-defining a set of high-quality data to guide the filtering process. Commonly, a classifier is employed to distinguish data between high and low quality \cite{gururangan2022whose, chowdhery2023palm, li2023textbooks}, which is dependent on the classifier's precision. 
\citeauthor{xie2024data} (\citeyear{xie2024data}) involve importance resampling to extract a subset from the raw dataset that closely aligns with the target domain distribution, while \citeauthor{engstrom2024dsdm} (\citeyear{engstrom2024dsdm}) propose using datamodels \cite{ilyas39datamodels} to assess data quality and optimize the model's performance.
We argue that these selection methods may be influenced by the bias towards "high-quality data". Additionally, they could potentially lack diversity in knowledge, a deficiency that can be mitigated by incorporating our knowledge coverage metric.

\paragraph{Knowledge analysis}
\citeauthor{kandpal2023large} (\citeyear{kandpal2023large}) uses salient entities within test questions and answers to assess the distribution of relevant knowledge in the training set, while \citeauthor{lucy2024aboutme} (\citeyear{lucy2024aboutme}) use the 'AboutMe' page to identify the main knowledge of a webpage. \citeauthor{lu2023instag} (\citeyear{lu2023instag}) involve annotating the tag fragments contained in each text of the instruction-tuning training set, but these tags are often viewed as topics rather than knowledge pieces. In contrast to these studies, we have developed a comprehensive, categorized collection of knowledge elements to tag and filter training data.


\section{Conclusion}
\label{sec:conclusion}


In this work, we propose a novel method for selecting high-quality data from a knowledge perspective. We create a comprehensive knowledge element pool covering various fields, 
and develop an effective method free from parameter updates for selecting high-knowledge data for both general and specific domains. Our extensive experimental results prove that our method outperforms all the baselines in knowledge-intensive and general understanding tasks, which suggests that the high-knowledge data we selected is of superior quality. In addition, our experiments also confirm that we can effectively improve the data quality in specific domains. Nevertheless, our work does have some limitations, which we will discuss in the Appendix A.

\clearpage

\section*{Acknowledgments}

This work was supported in part by the National Natural Science Foundation of China under Grant 62477001.

\bibliography{aaai25}










\appendix

\section{Discussion and limitation}
\label{sec:limitation}


\paragraph{More knowledge elements and better filtering}
In this study, we have constructed a knowledge element pool encompassing five domains, with a total size of 5M. First of all, we could develop a larger pool by extracting from a broader and more diverse range of sources. Additionally, we could refine the categorization of knowledge elements, for instance, by further subdividing the existing categories into secondary ones. Despite our efforts in knowledge filtering, we have noticed the presence of some truncated pieces of knowledge. This observation has inspired us to investigate a more meticulous process for filtering knowledge elements.




\paragraph{Knowledge metrics and scoring function}
The study utilized density and coverage as metrics to evaluate the knowledge contained within a text. More knowledge metrics could be explored in future work. It is also worth noting that the scoring formula was determined through experimental optimization. A theoretical validation of the scoring formula would contribute to its robustness.


\paragraph{Scaling up model and dataset}
Our research was constrained by hardware capabilities, which restricted our testing to models with approximately 1.1B parameters. This limitation has impeded our capacity to fully compare the efficacy of different methodologies. Expanding our experiments to include larger models, such as those with 7B or 13B parameters, and utilizing more extensive datasets would significantly enhance the scope and depth of our research.

\section{Social Impact}
\label{impact}

Large Language Models (LLMs) often exhibit a phenomenon known as hallucination, where they generate content that is not factually accurate, influenced by their training datasets \cite{Huang2023ASO}. This situation has a negative impact on our society, as it leads to issues like the creation of fake news, which raises concerns about the reliability of LLM.  To mitigate this issue, we propose the HKS method to select high-knowledge data, thereby enhancing the reliability and authenticity of the training data. Moreover, our approach efficiently selects data within a short timeframe and at reduced hardware costs. This not only boosts efficiency but also lowers carbon emissions from computational processes, contributing positively to sustainable development.

\section{GPT4 prompts}
\label{sec:GPT4_prompts}

For knowledge extraction and filtering, we guide GPT4 to collect a broader range of knowledge elements. Here we show the prompts.

\label{prompt_for_knowledge_extraction}
\begin{tcolorbox}[colback=white!95!gray,colframe=gray!50!black,rounded corners,label={prompt-dot1}, title={Prompts for knowledge extraction and filtering}]
\begin{lstlisting}[breaklines=true, xleftmargin=0pt, breakindent=0pt, columns=fullflexible, mathescape, numbers=none]
You are a language model training corpus annotator, your task is to identify the conceptual vocabulary in each text. You need to extract all the conceptual vocabulary from each text. 

The type of concept is mainly the concept in professional disciplines (such as: science, society, life, art, and culture). 

Requirements: 
Data annotation first states your identification reason and then gives your identification result. 

Constraints: 
First output the identification reason for each vocabulary, and then give the identified vocabulary. The prescribed output template is as follows: "XX is a concept in 'YY', which refers to... The corresponding identified vocabulary is: 'XX'". Where "XX" represents the identified concept vocabulary, and YY represents the professional discipline to which the concept vocabulary belongs. 
{{text}} 

If there are multiple concepts, please output multiple lines. Ensure that only one identification reason and concept vocabulary are output on each line. 

The text content is as follows: 
{{text}} 
\end{lstlisting}
\end{tcolorbox}

\label{prompt_for_knowldge_filter}
\begin{tcolorbox}[colback=white!95!gray,colframe=gray!50!black,rounded corners,label={prompt-dot2}, title={Knowledge filter}]
\begin{lstlisting}[breaklines=true, xleftmargin=0pt, breakindent=0pt, columns=fullflexible, mathescape, numbers=none]
You are a language model training corpus annotator, your task is to identify whether the given phrase is a conceptual vocabulary in the specified professional discipline. 

Requirements: 
Data annotation first states your identification reason and then gives your identification result. 

Input instructions: 
Input format: XX="...", YY="...", CC="...". Where "XX" represents the given word to be identified, YY represents the given professional discipline, and CC represents the context where XX is located. 

Constraints: 
1. First output the reason for vocabulary judgment, and then give the judgment result. The prescribed output template is as follows: "Because XX represents ..., and YY represents ..., so the judgment result: 'ZZ', and the confidence is ...". Where "XX" represents the given word to be identified, YY represents the given professional discipline, and ZZ represents the judgment result. 
2. The judgment result ZZ can only answer "yes" or "no". 

The text content is as follows: 
{XX="{{concept}}", YY="{{domain}}", CC="{{context}}"} 
\end{lstlisting}
\end{tcolorbox}

\section{Function Search Process}
\label{sec:function_search}



According to the method description, we need to search for a suitable function form for $f$ and $g$. From the definition 2 and 3, $d$ and $c$ take values between $[0, 1]$, so it is sufficient that both $f$ and $g$ are convex functions between $[0, 1]$. Thus we choose among the following functions for $f$ and $g$: $Identidy$, $sin(x)$ and $log(x+1)$ respectively. We randomly select 2k pairs from the training set, each pair consisting of two different texts. 

We recruit three annotators and guide the annotators according to the pre-defined instructions, and eventually average the results of all the annotators as the manual labeling results for each pair. We paid \$15/hour to each annotator. We clarified to the annotators the significance of their annotations on the study and assured them that our annotation task did not involve any implications for personal safety or mental health.
Subsequently, all the texts are scored through the scoring function. After score normalization, we calculate the softmax distribution of the score in each pair and compute the Spearman'rank correlation coefficient between the score distributions and manual labels. 

The results, as presented in Table \ref{formula_corr}, reveal that $d \cdot ln(c+1)$ exhibits the strongest correlation with the human-annotated results. Consequently, we select this formula as our final scoring function.


\label{a_instruction}
\begin{tcolorbox}[colback=white!95!gray,colframe=gray!50!black,rounded corners,label={prompt-dot3}, title={Instructions for human annotators}]
\begin{lstlisting}[breaklines=true, xleftmargin=0pt, breakindent=0pt, columns=fullflexible, mathescape, numbers=none]
You are a language model training corpus annotator. Give you two text samples, your task is to identify which sample contains more informative signal for pre-training a large-language model.

An informative data point should be well-formatted, contain some usable knowledge of the world, and strictly NOT have any harmful, racist, sexist, etc. content.

If you prefer the first one, give the response "0", otherwise the response "1".
\end{lstlisting}
\end{tcolorbox}

\begin{table*}[!ht]
    \centering
    \scalebox{0.85}{
    \begin{tabular}{lc|lc|lc}
    \toprule
        \textbf{Formula} & \textbf{Spearman cor.} & \textbf{Formula} & \textbf{Spearman cor.} & \textbf{Formula} & \textbf{Spearman cor.} \\ 
    \midrule
        $d \cdot c$ & 0.837 & $ln(d+1) \cdot c$ & 0.815 & $sin(d) \cdot c$ & 0.746 \\ 
        $d \cdot ln(c+1)$ & \textbf{0.855} & $ln(d+1) \cdot ln(c+1)$ & 0.834 & $sin(d) \cdot ln(c+1)$ & 0.760 \\ 
        $d \cdot sin(c)$ & 0.840 & $ln(d+1) \cdot sin(c)$ & 0.817 & $sin(d) \cdot sin(c)$ & 0.758 \\ 
    \bottomrule
    \end{tabular}
    }
    \caption{Correlation between scores of different scoring functions and human annotated labels.}
    \label{formula_corr}
\end{table*}


\section{Knowledge Element Pool}
\label{sec:knowledge_elements_pool}

\paragraph{Knowledge element categorization} We classify knowledge elements into five domains: \textit{science, society, culture, art, and life}. As it is challenging to label individual knowledge elements directly, we associate each element with a related passage, which serves as supplementary material for tagging. Specifically, when dealing with Wiki entries, we use the entry content that contains the corresponding knowledge element as supplementary text. For the OAG dataset, we use the abstract as the matching text. We have manually annotated 10,000 knowledge elements. During the annotation process, we use the supplementary text as the unit, identify the domain to which the text belongs, and tag the corresponding knowledge element to this domain. We employ BERT \cite{devlin-etal-2019-bert} as the base model to train the labeling model, designate 20\% of the annotated data as the test set, and achieve a labeling accuracy of 96.2\%.



\begin{figure}[ht]
    \centering
    \includegraphics[width=0.5\linewidth]{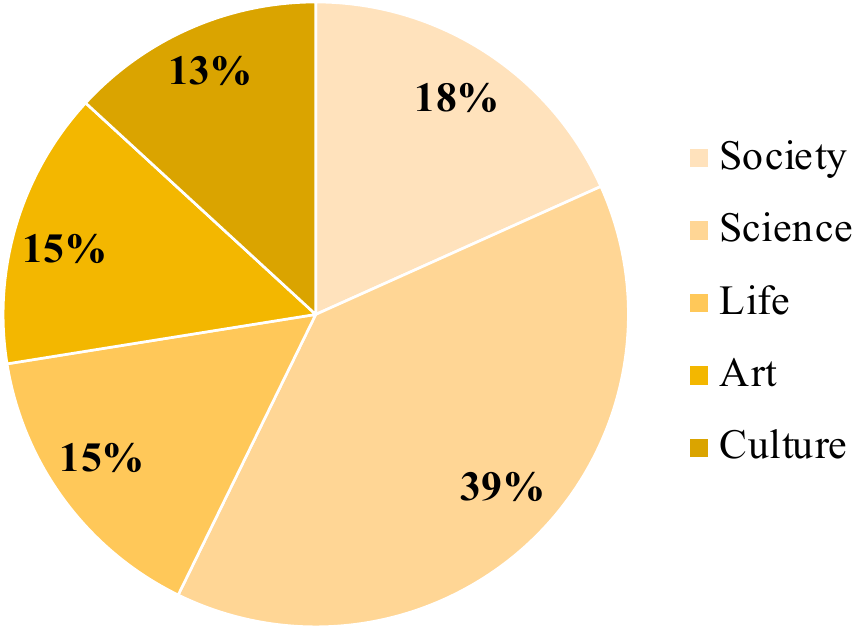}
    \caption{Knowledge elements distribution over different categories.}
    \vspace{0mm}
    \label{kc_distribution}
\end{figure}

\begin{figure}[!ht]
    \centering
    \begin{minipage}{1.0\linewidth}
        \centering
        \includegraphics[width=1.0\linewidth]{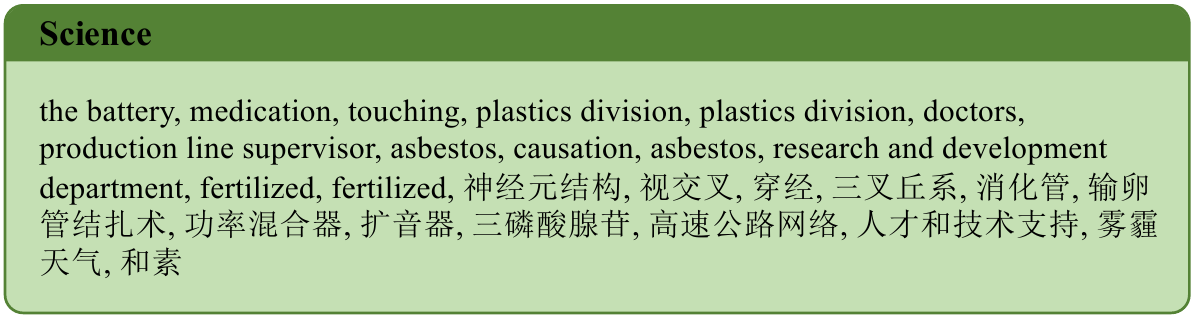}
        \vspace{0.01mm}
    \end{minipage}
    
    \begin{minipage}{1.0\linewidth}
        \centering
        \includegraphics[width=1.0\linewidth]{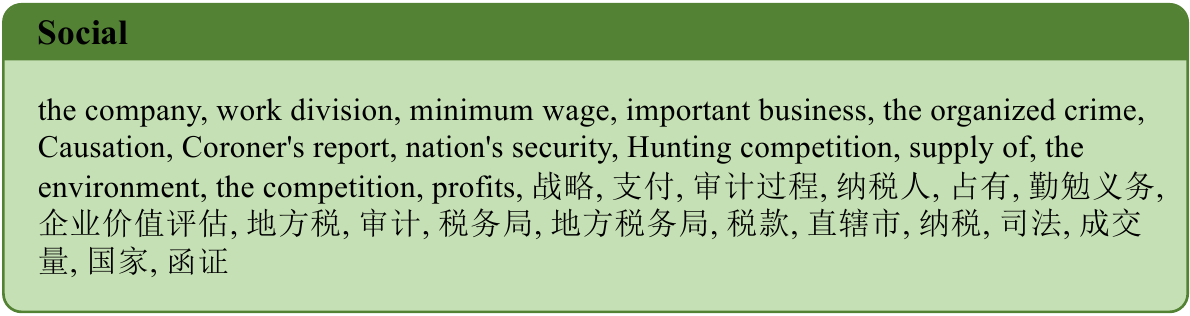}
        \vspace{0.01mm}
    \end{minipage}
    
    \begin{minipage}{1.0\linewidth}
        \centering
        \includegraphics[width=1.0\linewidth]{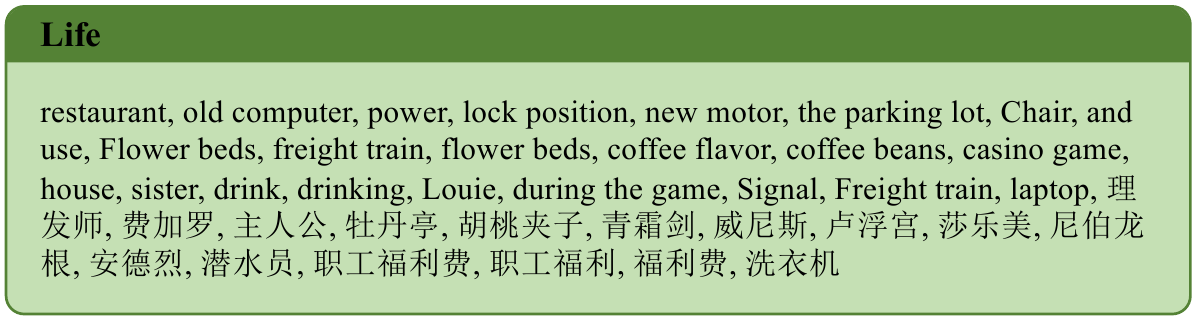}
    \end{minipage}
    \caption{Knowledge element cases over different categories.}
    \label{kc_cases}
\end{figure}

\paragraph{Knowledge element statistics} The number of knowledge elements directly derived from the entry titles and keywords account for 86.1\% of the entire pool, significantly higher than those extracted by GPT-4. Given that this part is mostly written by numerous experts, we can ensure the high quality of knowledge elements. For the knowledge element pool we have built up, we have counted the distribution of knowledge in each domain and the results are shown in Figure \ref{kc_distribution}. In addition, we give examples of the knowledge of some of the domains in Table \ref{kc_cases}.

\paragraph{Domain-specific knowledge density and coverage}

We give a formal definition of knowledge density and coverage for a target domain $m$:

\begin{definition}
    \label{def:df}
    Given a text sample $x$, $n_{km}$ is the total number of knowledge elements that belong to domain $m$ in the sample, and $n_p$ is the text token length, \textbf{Knowledge Density for Domain $m$ ($d_m$)} is defined as $d_m(x) = n_{km} / n_p$.
\end{definition}

\begin{definition}
    \label{def:cf}
    Given a text sample $x$, $\widetilde{n}_{km}$ is the total number of non-duplicated knowledge elements that belong to domain $m$ in the sample, and $N_{km}$ the total number in the domain $m$. \textbf{Knowledge Coverage for Domain $m$ ($c_m$)} is defined as $c_m(x) = \widetilde{n}_{km} / N_{km}$.
\end{definition}

\section{Model training details}
\label{sec:model_training}

We train autoregressive decoder-only Transformer models with a standard language modeling objective. Our language models follow the Bloom-style architecture, the model structure is given in Table \ref{tab:model_structure}. While training our models, we use Adam with $\beta_1 = 0.9$ and $\beta_2 = 0.95$, We utilize a cosine learning rate scheduler with a learning rate of $2.5 \times 10^{-4}$, and a batch size of 2048. The models use a context window length of 2048.

\begin{table*}[t]
    \centering
    \renewcommand\arraystretch{1.5}
    \begin{tabular}{lr|lr|lr}
        \toprule
            \textbf{Hyperparameter} & \textbf{Value} & \textbf{Hyperparameter} & \textbf{Value} & \textbf{Hyperparameter} & \textbf{Value} \\
        \midrule
             Precision & float16 & Attention heads & 16 & Activation & GELU\\
             Layers & 24 & Vocab size & 250,680 & Position embedding & Alibi \\
             Hidden dimension & 1536 & Sequence length & 2048 & Tied embedding & True\\
        \bottomrule
    \end{tabular}
    \caption{The hyperparameters of model structure.}
    \label{tab:model_structure}
\end{table*}

\section{Detailed Experiment Results}
\label{sec:model_evaluation}

We evaluate our models across a total of 16 sub-tasks. For each task, we utilize 5 random samples from the training set as demonstrations. For Chinese general understanding tasks, we select several tasks from CLUE \cite{xu2020clue} and FewClue \cite{xu2021fewclue} benchmark: CSLDCP(Chinese Science Literature discipline Classification), WSC(The Winograd Schema Challenge, Chinese Version), AFQMC(The Ant Financial Question Matching Corpus), CSL(Chinese Scientific Literature), OCNLI(Original Chinese Natural Language Inference), NCPT(NER-Clue-Per-Type), ChID(Chinese IDiom cloze test), NCEE(NER-Clue-Extract-Entity), DRCD(Delta Reading Comprehension Dataset).

We divide the test benchmarks into four categories.
\begin{itemize}[leftmargin=*]
    \item English knowledge-intensive tasks: ARC-C, OpenBookQA, MMLU
    \item English general understanding tasks: BBH, RTE, WIC, COPA, BoolQ
    \item Chinese knowledge-intensive tasks: CMMLU, C-Eval
    \item Chinese general understanding tasks: CSLDCP, WSC, AFQMC, CSL, OCNLI, NCPT, ChID, NCEE, DRCD
\end{itemize}

Table \ref{main_results_detailed} shows the detailed few-shot results of downstream tasks for main results, and Table \ref{sampling_results_detailed} shows the detailed results for the experiment of comparison between top-$k$ and sampling.

\begin{table*}[!ht]
    \centering
    \begin{tabular}{l|cccc|cccc}
    \toprule
        \textbf{Method} & \textbf{Random} & \textbf{PPL} & \textbf{EL2N} & \textbf{DSIR} & \textbf{$d$} & \textbf{$c$} & \textbf{HKS inverse} & \textbf{HKS} \\ 
    \midrule
        ARC-C & 19.37 & 23.81 & 24.15 & 21.84 & 21.50 & 23.38 & 20.05 & 23.55 \\ 
        OpenBookQA & 26.40 & 24.00 & 24.60 & 11.20 & 19.00 & 27.60 & 16.20 & 27.50 \\ 
        MMLU & 25.53 & 25.52 & 24.81 & 24.73 & 24.65 & 24.82 & 24.62 & 27.91 \\ 
        CMMLU & 25.34 & 26.00 & 25.62 & 25.49 & 25.19 & 25.60 & 25.67 & 27.85 \\ 
        C-Eval & 24.96 & 22.92 & 26.61 & 25.72 & 24.39 & 26.82 & 27.42 & 26.89 \\ 
        BBH & 27.80 & 28.34 & 28.40 & 27.83 & 28.51 & 27.51 & 27.86 & 29.66 \\ 
        RTE & 48.50 & 47.00 & 49.50 & 46.00 & 52.25 & 50.50 & 48.00 & 50.25 \\ 
        WIC & 50.00 & 52.75 & 50.75 & 50.75 & 51.50 & 49.75 & 50.75 & 49.75 \\ 
        COPA & 55.00 & 39.00 & 55.00 & 54.00 & 45.00 & 46.00 & 55.00 & 55.00 \\ 
        BoolQ & 56.50 & 62.50 & 57.50 & 62.00 & 41.50 & 41.50 & 49.00 & 60.00 \\ 
        CSLDCP & 4.50 & 0.00 & 3.00 & 5.00 & 5.50 & 5.00 & 1.50 & 14.00 \\ 
        WSC & 60.50 & 41.00 & 62.00 & 62.00 & 62.00 & 62.00 & 50.00 & 62.00 \\ 
        AFQMC & 47.25 & 46.75 & 49.50 & 46.50 & 50.50 & 49.50 & 49.75 & 49.75 \\ 
        CSL & 48.00 & 51.00 & 48.00 & 50.00 & 46.00 & 49.00 & 47.00 & 47.50 \\ 
        OCNLI & 31.00 & 32.50 & 31.00 & 34.50 & 31.50 & 32.50 & 34.50 & 34.00 \\ 
        NCPT & 27.00 & 22.00 & 12.50 & 11.50 & 5.50 & 24.00 & 17.00 & 23.50 \\ 
        ChID & 12.00 & 9.00 & 11.00 & 3.00 & 5.00 & 8.00 & 12.00 & 13.00 \\ 
        NCEE & 6.00 & 2.50 & 2.00 & 2.50 & 10.00 & 10.50 & 9.00 & 17.50 \\ 
        DRCD & 21.70 & 2.13 & 0.06 & 0.69 & 3.54 & 6.52 & 6.14 & 26.74 \\
    \bottomrule
    \end{tabular}
    \caption{Detailed downstream task results for main tasks, averaged over 3 seeds.}
    \label{main_results_detailed}
\end{table*}

\begin{table*}[!ht]
    \centering
    \begin{tabular}{l|cc|cc|cc}
    \toprule
        \multirow{2}{*}{\textbf{Method}} & \multicolumn{2}{c|}{\textbf{$d$}} & \multicolumn{2}{c|}{\textbf{$c$}} & \multicolumn{2}{c}{\textbf{HKS}} \\
    \cmidrule{2-7}
        ~ & \textbf{top-$k$} & \textbf{sampling} & \textbf{top-$k$} & \textbf{sampling} & \textbf{top-$k$} & \textbf{sampling} \\
    \midrule
        ARC-C & 21.50 & 20.14 & 23.38 & 22.36 & 23.55 & 23.72 \\ 
        OpenBook QA & 19.00 & 27.80 & 27.60 & 24.00 & 27.50 & 24.80 \\ 
        MMLU & 24.65 & 25.09 & 24.82 & 24.74 & 27.91 & 26.14 \\ 
        CMMLU & 25.19 & 25.57 & 25.60 & 25.37 & 27.85 & 26.65 \\ 
        C-Eval & 24.39 & 25.34 & 26.82 & 28.12 & 26.89 & 26.82 \\ 
        BBH & 28.51 & 28.47 & 27.51 & 26.66 & 29.66 & 28.53 \\ 
        RTE & 52.25 & 50.75 & 50.50 & 48.25 & 50.25 & 48.25 \\ 
        WIC & 51.50 & 50.00 & 49.75 & 50.00 & 49.75 & 49.75 \\ 
        COPA & 45.00 & 55.00 & 46.00 & 42.00 & 55.00 & 50.00 \\ 
        BoolQ & 41.50 & 56.00 & 41.50 & 48.50 & 60.00 & 41.50 \\ 
        CSLDCP & 5.50 & 4.00 & 5.00 & 7.00 & 14.00 & 2.50 \\ 
        WSC & 62.00 & 61.50 & 62.00 & 62.00 & 62.00 & 62.00 \\ 
        AFQMC & 50.50 & 46.50 & 49.50 & 51.50 & 49.75 & 45.20 \\ 
        CSL & 46.00 & 52.00 & 49.00 & 48.00 & 47.50 & 47.00 \\ 
        OCNLI & 31.50 & 36.00 & 32.50 & 34.00 & 34.00 & 33.50 \\ 
        NCPT & 5.50 & 21.50 & 24.00 & 20.50 & 23.50 & 32.00 \\ 
        ChID & 5.00 & 6.00 & 8.00 & 8.50 & 13.00 & 7.50 \\ 
        NCEE & 10.00 & 6.50 & 10.50 & 5.00 & 17.50 & 5.50 \\ 
        DRCD & 3.54 & 3.07 & 6.52 & 3.86 & 26.74 & 7.66 \\
    \bottomrule
    \end{tabular}
    \caption{Detailed downstream task results for top-$k$ and sampling, averaged over 3 seeds.}
    \label{sampling_results_detailed}
\end{table*}

\section{Document cases}
\label{sec:document_cases}

\begin{figure*}[!ht]
    \centering   
    \includegraphics[width=0.7\linewidth,height=0.9\textheight,keepaspectratio]{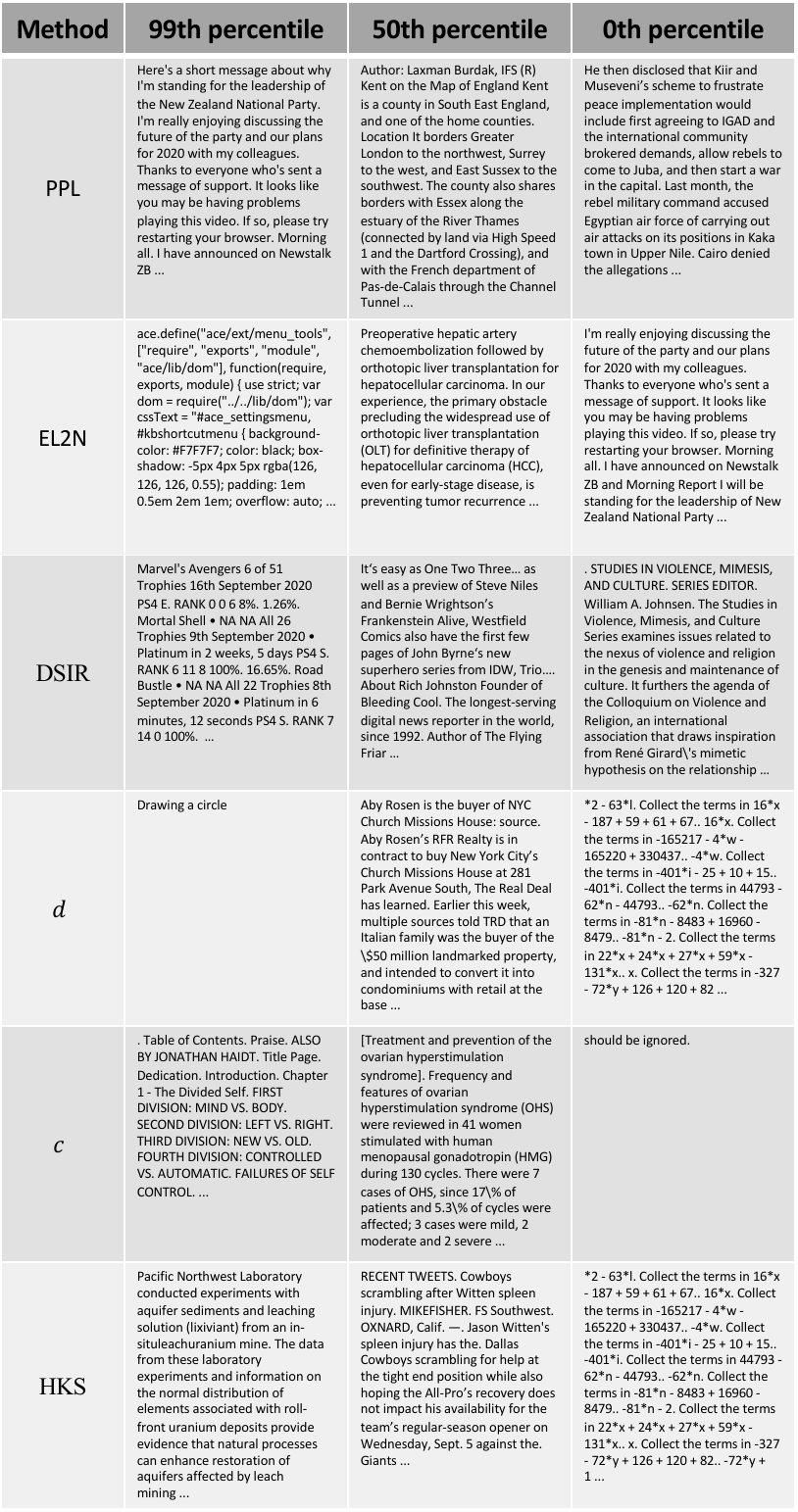}
    \caption{Data selection cases on Pile.}
    \label{ds_cases}
\end{figure*}

\begin{figure*}[!ht]
    \centering
    \includegraphics[width=0.7\linewidth,height=0.9\textheight,keepaspectratio]{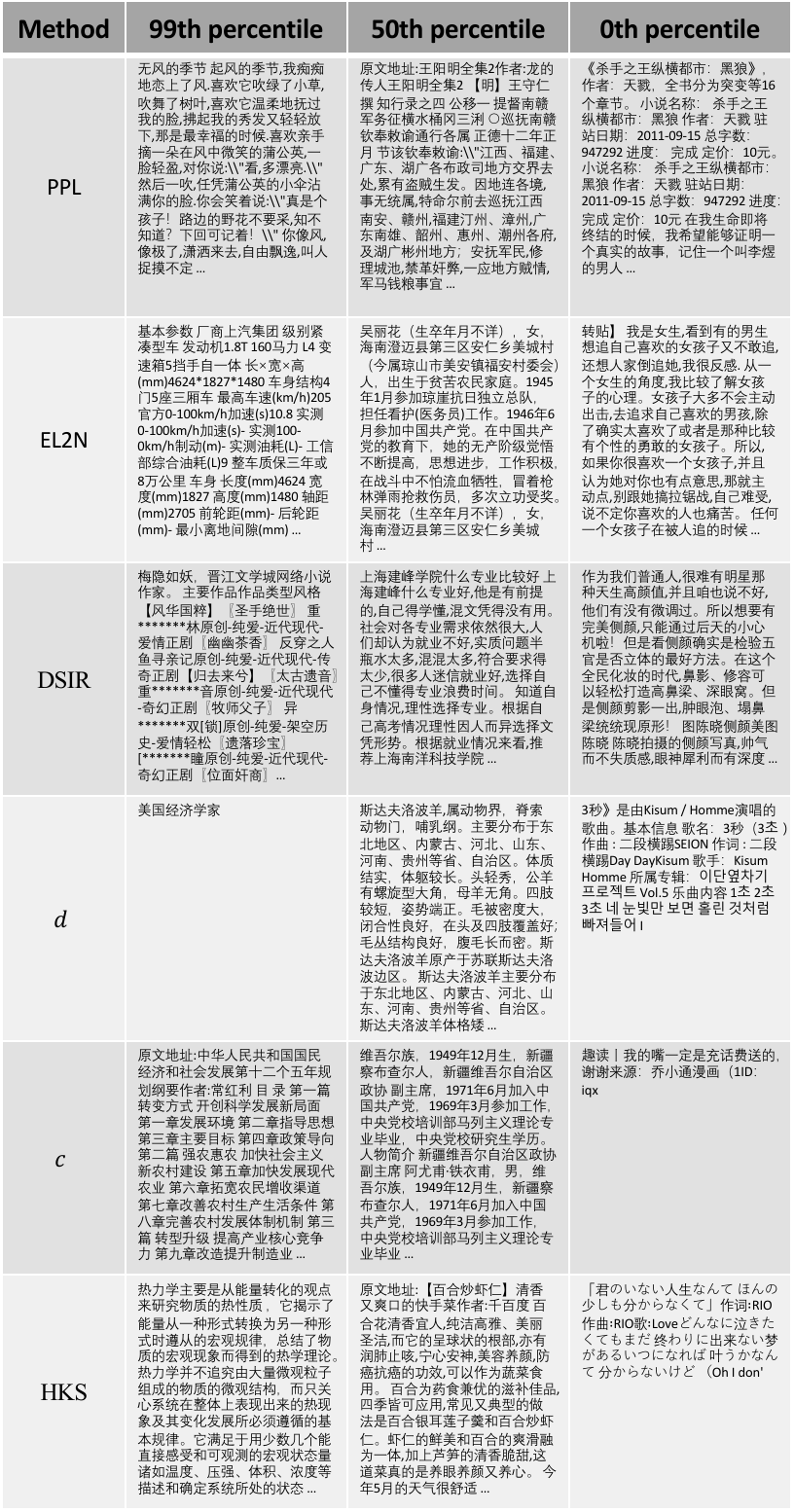}
    \caption{Data selection cases on Wudao.}
    \label{ds_cases_wudao}
\end{figure*}

We present cases from the raw documents of both the Pile datasets. These documents are selected from the 0th, 50th, and 99th percentiles of various score ratings, as depicted in Figure \ref{ds_cases} and \ref{ds_cases_wudao}.

\end{document}